\documentclass[acmsmall]{acmart}
\usepackage{multirow}
\usepackage{array}
\usepackage{subfigure}
\usepackage{caption}
\usepackage{graphicx}
\usepackage{hyperref}
\usepackage{url}

\AtBeginDocument{%
  \providecommand\BibTeX{{%
    \normalfont B\kern-0.5em{\scshape i\kern-0.25em b}\kern-0.8em\TeX}}}

\setcopyright{acmlicensed}
\copyrightyear{2023}
\acmYear{2023}
\acmDOI{XXXXXXX.XXXXXXX}

\acmJournal{JACM}
\acmVolume{37}
\acmNumber{4}
\acmArticle{111}
\acmMonth{8}

\begin{document}

\title{Two Heads Are Better Than One: Integrating Knowledge from Knowledge Graphs and Large Language Models for Entity Alignment}

%
% The "author" command and its associated commands are used to define
% the authors and their affiliations.
% Of note is the shared affiliation of the first two authors, and the
% "authornote" and "authornotemark" commands
% used to denote shared contribution to the research.
%\author{Linyao Yang}
%\email{yangly@zhejianglab.com}
%\orcid{0000-0003-0826-9453}
%\email{webmaster@marysville-ohio.com}
%\affiliation{%
%  \institution{Institute for Clarity in Documentation}
%  \streetaddress{P.O. Box 1212}
%  \country{USA}

\author{Linyao Yang}
\email{yangly@zhejianglab.com}
\orcid{0000-0003-0826-9453}

\author{Hongyang Chen}
\authornote{Corresponding author.}
\email{dr.h.chen@ieee.org; hongyang@zhejianglab.com}
\affiliation{%
  \institution{Zhejiang Lab}
%  \streetaddress{P.O. Box 1212}
  \city{Hangzhou}
  \state{Zhejiang}
  \country{China}
  \postcode{311121}
}

\author{Xiao Wang}
\affiliation{%
  \institution{Anhui University}
%  \streetaddress{1 Th{\o}rv{\"a}ld Circle}
  \city{Hefei}
  \state{Anhui}
  \country{China}
  \postcode{230039}}
\email{xiao.wang@ahu.edu.cn}

\author{Jing Yang}
\email{yangjing2020@ia.ac.cn}

\author{Fei-Yue Wang}
\email{feiyue.wang@ia.ac.cn}
\affiliation{%
  \institution{Institute of Automation, Chinese Academy of Sciences}
%  \streetaddress{P.O. Box 1212}
  \city{Beijing}
%  \state{Zhejiang}
  \country{China}
  \postcode{100190}
}

\author{Han Liu}
\email{liu.han.dut@gmail.com}
\affiliation{%
  \institution{Dalian University of Technology}
%  \streetaddress{30 Shuangqing Rd}
  \city{Dalian}
  \state{Liaoning}
  \country{China}
  \postcode{116081}}

\renewcommand{\shortauthors}{Yang and Chen, et al.}

\begin{abstract}

Entity alignment, which is a prerequisite for creating a more comprehensive Knowledge Graph (KG), involves pinpointing equivalent entities across disparate KGs. Contemporary methods for entity alignment have predominantly utilized knowledge embedding models to procure entity embeddings that encapsulate various similarities—structural, relational, and attributive. These embeddings are then integrated through attention-based information fusion mechanisms. Despite this progress, effectively harnessing multifaceted information remains challenging due to inherent heterogeneity. Moreover, while Large Language Models (LLMs) have exhibited exceptional performance across diverse downstream tasks by implicitly capturing entity semantics, this implicit knowledge has yet to be exploited for entity alignment. In this study, we propose a Large Language Model-enhanced Entity Alignment framework (LLMEA), integrating structural knowledge from KGs with semantic knowledge from LLMs to enhance entity alignment. Specifically, LLMEA identifies candidate alignments for a given entity by considering both embedding similarities between entities across KGs and edit distances to a virtual equivalent entity. It then engages an LLM iteratively, posing multiple multi-choice questions to draw upon the LLM's inference capability. The final prediction of the equivalent entity is derived from the LLM's output. Experiments conducted on three public datasets reveal that LLMEA surpasses leading baseline models. Additional ablation studies underscore the efficacy of our proposed framework.
%  Entity alignment that identifies identical entities across different knowledge graphs (KGs) is a prerequisite for building a more complete KG. Recent entity alignment methods focus on applying knowledge embedding models to learn entity embeddings capturing different similarities, such as structural, relational, and attribute, and fusing different embeddings with attentive information fusion modules. However, it is not trivial to leverage diverse information in EA because of the heterogeneity. Moreover, the great success of large language models (LLMs) in various downstream tasks shows their powerful ability to implicitly model entity semantics. However, this implicit knowledge contained in LLMs have not been exploited in entity alignment. In this paper, we propose LLMEA, a large language model enhanced entity alignment framework, to fuse the structural knowledge from KGs and the semantic knowledge from LLMs to boost the performance of entity alignment. Specifically, for a target entity to be aligned, it first obtains candidate entities based on the structural embedding similarity and the edit distance from a virtual equivalent entity. Then, it iteratively constructs multiple multi-choice question tasks to the LLM to levarage its knowledge and reasoning ability. Finally, the equivalent entity can be predicted from LLM's output. Extensive experimental results on three public datasets show that LLMEA outperforms state-of-the-art baselines and ablation studies demonstrate the effectiveness of our proposed framework.
\end{abstract}


%\ccsdesc{Computing methodologies~Natural language processing}


\received{30 January 2024}
\received[revised]{XX XX 2024}
\received[accepted]{XX XX 2024}

\maketitle

\section{Introduction}

Knowledge Graphs (KGs) encapsulate real-world knowledge in the form of triples, resulting in heterogeneous graphs where concepts are represented as nodes and the relationships between these concepts are represented as edges \cite{UrbanKG}. In recent years, the proliferation of KGs has been remarkable, with various KGs such as DBpedia \cite{DBpedia} and YAGO \cite{YAGO} being developed. KGs inherently present the overarching structure of knowledge and reasoning chains in a readable format, making them applicable for knowledge modeling. Consequently, KGs have been effectively leveraged in a wide array of knowledge-centric applications, such as few-shot learning \cite{3510030}, recommendation systems \cite{2926718,3637216,3635273,3641288}, and drug discovery \cite{lin2020kgnn}.

%Knowledge graphs (KGs) represent real-world knowledge into triples, forming heterogeous graphs, in which concepts are represented as nodes while the relations among concepts are represented as edges. Recent years have witnessed the proliferation of KGs, various KGs, such as DBpedia \cite{DBpedia} and YAGO \cite{YAGO}, have been constructed. KGs intuitively display the overall structure of knowledge and reasoning chains in a readable format, making them an ideal choice for knowledge modeling. As a result, KGs have been successfully applied to diverse knowledge-driven applications, such as recommendation system \cite{2926718}, question answering \cite{3539289}, and drug discovery \cite{lin2020kgnn}.

\begin{figure}[h]
  \centering
  \includegraphics[width=1.0\linewidth]{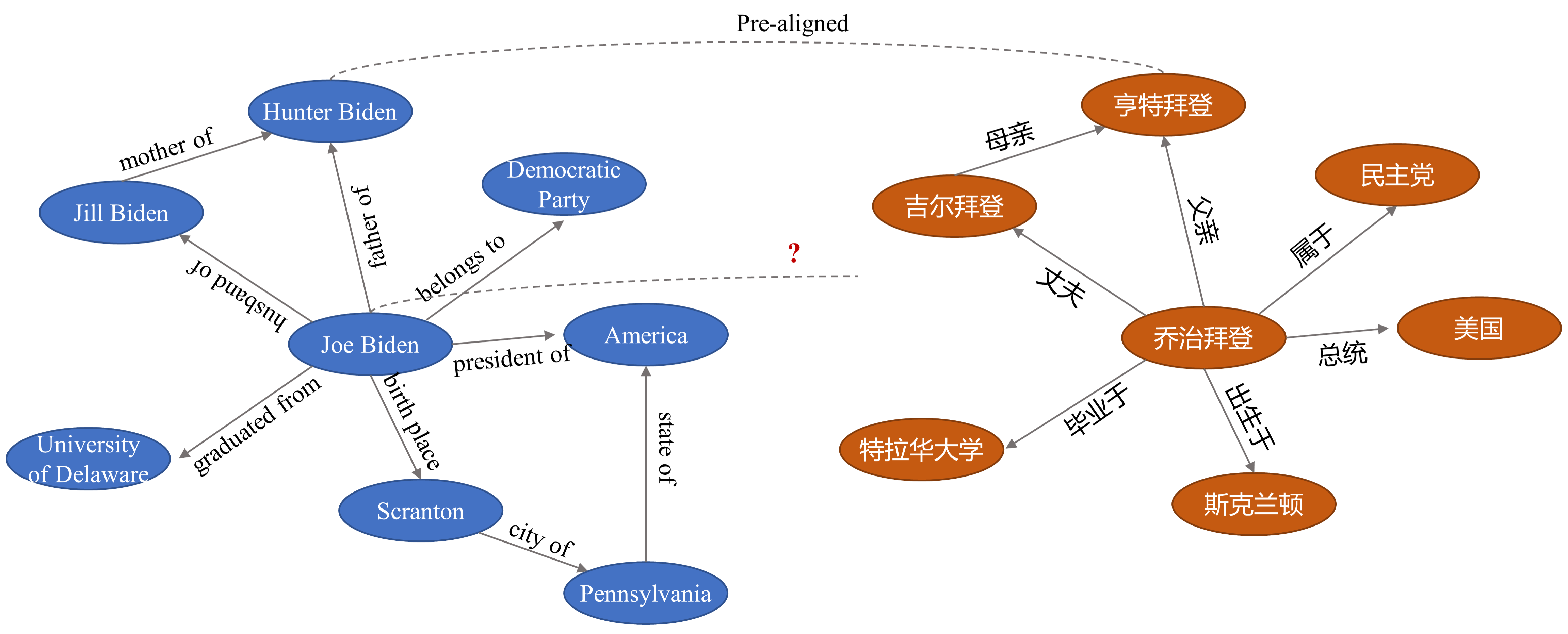}
  \caption{An illustration of entity alignment between an English KG and a Chinese KG.}
%  \Description{A woman and a girl in white dresses sit in an open car.}
\label{fig:illustrate}
\end{figure}

In practical settings, a KG is typically constructed from a solitary data source. Owing to the expansive nature of worldly knowledge, achieving complete coverage for one single KG remains improbable. Meanwhile, multiple different KGs may store extra or complementary knowledge. For instance, general KGs usually only contain some common encyclopedia knowledge, whereas more detailed domain knowledge can be found in specific domain KGs. Consequently, entity alignment that aims at identifying equivalent entities across KGs to integrate multiple KGs into a consolidated one has garnered substantial research attention. Fig.~\ref{fig:illustrate} illustrates an example of entity alignment. Typically, it involves computing the similarities between entities across KGs. Most recent entity alignment methods \cite{MTransE,GCN-Align,JAPE} resort to knowledge embedding models or graph neural networks \cite{3633518,3495161,HGATE} to learn embeddings capturing entities' similarities. To capture different aspects of proximities between entities, diverse embeddings are learned based on different kinds of information such as structure, name, and relation, which are then attentively fused to learn the comprehensive similarities between entities \cite{CEA,DAT,XGEA}. Most of these information fusion methods are early fusion \cite{zhang2021deep} methods, which combines information captured from multiple aspects into a low-dimensional space. However, different aspects of information have heterogeneous features, making it not a trival task to align their semantics and effectively fuse their complementary information. 

% More specifically, they can implicitly model relationships between entities based on co-occurrence frequencies in the corpus.

%In recent years, large language models (LLMs) have demonstrated outstanding performance across various natural language processing tasks, such as question answering \cite{QA-GNN}, information extraction \cite{3358119}, and text generation \cite{pascual-etal-2021-plug-play}. With the incease of model size, recent LLMs such as ChatGPT and GPT-4 have exhibited emergent abilities, displaying significant potential in diverse applications such as code generation \cite{3417058}, recommendation \cite{3463069}, and game play \cite{meta2022human}. Additionally, studies have revealed that LLMs can effectively memorize some facts and knowledge cembedded in the training corpus. As the example demonstrated in Fig.~\ref{fig:llm-know}, LLMs pretrained on enormous texts from different sources contain rich knowledge about real-world entities, which implicitly capture similarities between entities across KGs. Based on this knowledge, they are able to distinguish entities that are actually the same but have different surface names. Consequently, LLMs can be employed to identify equivalent entities by leveraging the semantic knowledge encoded in their extensive parameters, which remains unexplored in the existing literature.

\begin{figure}[h]
  \centering
  \includegraphics[width=1.0\linewidth]{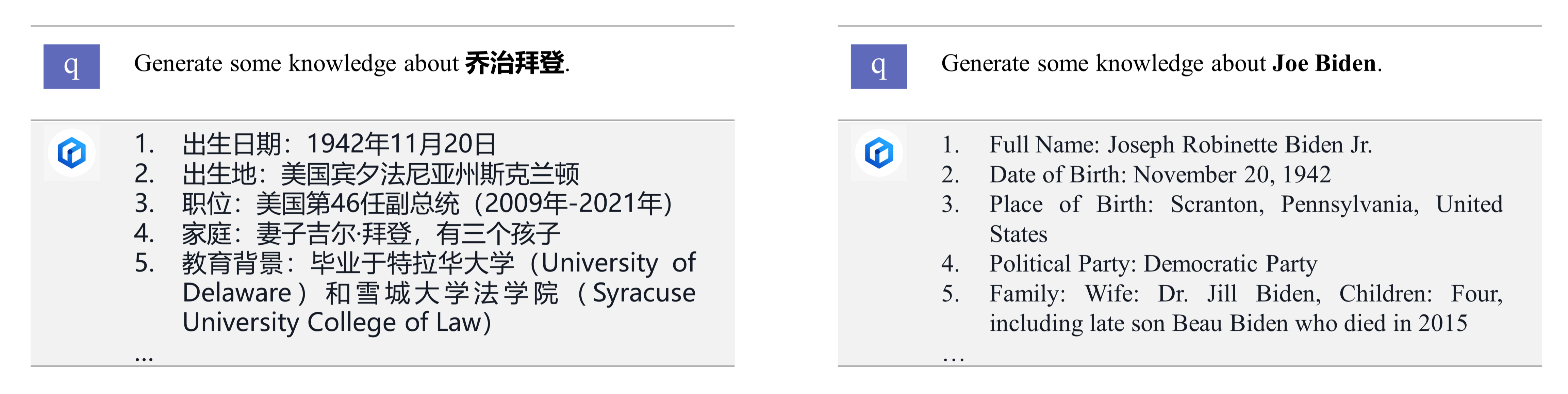}
  \caption{An illustration about LLMs modeling semantic knowledge of entities.}
%  \Description{A woman and a girl in white dresses sit in an open car.}
\label{fig:llm-know}
\end{figure}

In recent years, Large Language Models (LLMs) have showcased remarkable performance across various natural language processing applications, including question answering \cite{QA-GNN}, social network analysis \cite{zeng2024large}, and text generation \cite{pascual-etal-2021-plug-play}. With the escalation in model size, recent LLMs like ChatGPT have demonstrated emergent capabilities, revealing significant potential across diverse applications such as code generation \cite{3417058}, recommendation \cite{3463069}, and gameplay \cite{meta2022human}. Moreover, research has unveiled that LLMs possess the ability to effectively retain certain facts and knowledge embedded within the training corpus. As illustrated in the example depicted in Fig.~\ref{fig:llm-know}, LLMs pretrained on vast corpora from various sources encapsulate extensive knowledge about real-world entities, implicitly capturing similarities between entities across KGs. Drawing upon this knowledge, they can discern entities that represent the same concept but possess different surface names. Consequently, LLMs can be leveraged to identify equivalent entities, an avenue that remains largely unexplored in the existing literature.

% Additionally, LLMs and KGs cannot be directly integrated due to their distinct characteristics.
To integrate semantic knowledge from both KGs and LLMs, we propose a novel framework termed Large Language Model enhanced Entity Alignment framework (LLMEA). Conventional methods often rely on calculating entity name embeddings based on word embeddings from LLMs to utilize the knowledge from LLMs. However, this approach is limited to word-level similarity and fails to harness the profound semantics and inference capabilities inherent in LLMs. The recent advancement in prompt engineering techniques offers a promising avenue for interacting with LLMs effectively. Hence, a straightforward approach involves framing entity alignment as prompts to leverage the wealth of knowledge encoded within LLMs. For instance, we can task LLMs with discerning the equivalent entity to "Joe Biden" among all entities within a Chinese KG. However, this approach not only disregards the crucial structural knowledge within KGs but also proves insufficient for tackling large-scale entity alignment challenges due to the inherent input token size limitations of LLMs. Furthermore, inundating LLMs with too many options can potentially distort their judgment, leading to erroneous outcomes. 

To address these limitations, we introduce a method to filter candidate alignment entities for the target entity based on the structural features of KGs and the internal knowledge of LLMs. Specifically, we employ a Relation-Aware Graph ATtention network (RAGAT) to learn entity embeddings that capture structural and relational similarities. Subsequently, candidate equivalent entities are identified based on the learned embedding similarity. Simultaneously, we utilize an LLM to generate a virtual equivalent entity by leveraging its internal knowledge through a carefully designed knowledgeable prompt. Additional candidate entities are then identified based on the edit distance to the virtual equivalent entity. Ultimately, these candidate entities are presented as options to form multi-choice questions, which are input into the LLM to predict the final alignment result. Furthermore, we decompose the multi-choice question task into multiple rounds of choices to mitigate confusion and lengthy chain of thought arising from an excessive number of options. LLMEA can be regarded as a novel late fusion method of knowledge from KGs and LLMs. 

The primary contributions of this paper are summarized as follows: 
\begin{itemize}
\item We propose the LLMEA framework for entity alignment, which effectively fuses the knowledge from KGs and LLMs and employs the exceptional inference ability of LLMs. To the best of our knowledge, we are the first to exploit LLMs to enhance entity alignment.
\item We design a novel prompt method to leverage the internal knowledge of LLMs for filtering candidate entities. Additionally, we propose an entity alignment prediction method based on multi-choice questions, providing a distinctive approach to this task.
\item Experimetal results show that the proposed LLMEA framework achieves state-of-the-art performance across three datasets. These results serve as compelling evidence of the framework's effectiveness in advancing entity alignment methodologies.
\end{itemize}

%(1) We propose the LLMEA framework for entity alignment, which effectively fuses the knowledge from KGs and LLMs and employs the exceptional inference ability of LLMs. To the best of our knowledge, we are the first to exploit LLMs to enhance entity alignment. (2) We design a novel prompt method to leverage the internal knowledge of LLMs for filtering candidate entities. Additionally, we propose an entity alignment prediction method based on multi-choice questions, providing a distinctive approach to this task. (3) Experimetal results show that the proposed LLMEA framework achieves state-of-the-art performance across three datasets. These results serve as compelling evidence of the framework's effectiveness in advancing entity alignment methodologies.

The remainder of this paper is organized as follows: Section 2 provides an overview of existing trends and techniques for entity alignment and LLMs. Section 3 introduces the details of the proposed LLMEA framework. Following that, Section 4 presents the empirical evaluation. Lastly, Section 5 summarizes the main conclusions of this paper.

%The remainder of this paper is structured as follows. Section 2 gives an overview about existing trends and techniques for entity alignment and LLM. Section 3 introduces details of the proposed LLMEA framework. Then, section 4 gives the empirical evaluation. Lastly, section 5 summarizes the main conclusions of this paper.

\section{Related Work}

In this section, we present relevant efforts in entity alignment and highlight recent advances in LLMs, with a particular focus on their knowledgeable applications.

\subsection{Entity Alignment}

Entity alignment endeavors to identify entities sharing the same real-world identity across disparate KGs, which is a critical process for amalgamating knowledge from multiple KGs. This endeavor holds potential to enhance various downstream KG applications, including knowledge graph completion \cite{HackRL} and entity classification \cite{yang2021learning}. Early attempts at entity alignment primarily rely on crowdsourcing or manually crafted features, which are labor-intensive and inefficient. In recent years, with the advent of representation learning, embedding-based techniques have emerged as the predominant approach for entity alignment \cite{zhao2020experimental}. These methods typically represent entities as low-dimensional vectors based on a limited number of pre-aligned alignment seeds, effectively capturing semantic proximities among entities and facilitating scalability to large-scale KGs.

%Entity alignment aims to identify entities with the same real-world identity from different KGs, which is crucial for the knowledge fusion of multiple KGs and may benefit various downstream KG applications such as knowledge graph completion \cite{HackRL} and entity classification \cite{yang2021learning}. Early attempts at entity alignment are mostly based on crowdsourcing \cite{HIKE} or human-crafted features \cite{Big-Align}, which are labor-intensive and inefficient. In recent years, with the development of representation learning, embedding-based methods have emerged as the mainstream for entity alignment \cite{zhao2020experimental}. Embedding-based entity alignment methods typically represent entities as low-dimensional vectors based on a few pre-aligned alignment seeds, which effectively capture entities' semantic proximities and are easier to extend to large-scale KGs.

Some subset embedding-based methods exclusively utilize structural embeddings to measure entity similarity across KGs. For instance, MTransE \cite{MTransE} employs TransE \cite{TransE} to learn structural embeddings from triples. IPTransE \cite{IPTransE} iteratively refines TransE embeddings by considering highly confident aligned entities. BootEA \cite{BootEA} proposes a bootstrapping approach, iteratively labeling likely equivalent entities as training data for learning alignment-oriented KG embeddings. GCN-Align \cite{GCN-Align} is the first to utilize Graph Convolutional Networks (GCNs) to learn entity embeddings, thereby representing different KGs in a unified vector space through parameter sharing. KECG \cite{KECG} extends the Graph Attention Network (GAT) to ignore unimportant neighbors for alignment via an attention mechanism. AliNet \cite{AliNet} introduces a novel GCN that aggregates both direct and distant neighborhood information while learning entity representation, aiming to mitigate non-isomorphism in neighborhood structures. RREA \cite{RREA} designs a relation reflection transformation operation to acquire relation-specific embeddings for each entity, constraining the transformation matrix to be orthogonal to maintain unchanged norms and relative distances.

%Some subset embedding-based methods exclusively leverage structural embeddings to calculate the similarity between entities across KGs. For instance, MTransE \cite{MTransE} employs TransE \cite{TransE} to learn structural embeddings from triples. IPTransE \cite{IPTransE} iteratively refines TransE embeddings by considering high-confidence aligned entities. BootEA \cite{BootEA} proposes a bootstrapping approach, iteratively labeling likely equivalent entities as training data for learning aignment-oriented KG embeddings. GCN-Align \cite{GCN-Align} firstly adopts Graph Convolutional Networks (GCNs) to learn entity embeddings, which represents different KGs into a unified vector space through parameter sharing. KECG \cite{KECG} extends the Graph ATtention network (GAT) to ignore unimportant neighbors for alignment through attention mechanism. AliNet \cite{AliNet} introduces a novel GCN that aggregates both direct and distant neighborhood information while learning entity representation, aiming to mitigate non-isomorphism in neighborhood structures. RREA \cite{RREA} designs a relation reflection transformation operation to obtain relation specific embeddings for each entity, restricting the transformation matrix to be orthogonal to maintain unchanged norms and relative distances.

Several other embedding-based entity alignment methods propose the acquisition of embeddings that encapsulate diverse facets of proximities, encompassing attributes, semantic information from names, and other auxiliary information. These distinct embeddings are subsequently combined to assess the similarity between entities. For instance, JAPE \cite{JAPE} introduces a unified embedding model that preserves attributes jointly, embedding the structures of two KGs into a unified vector space and refining it by exploiting attribute correlations within the KGs. JarKA \cite{JarKA} introduces a joint framework that consolidates alignments deduced from both attributes and structures. CEA \cite{CEA} employs three key features—structural, semantic, and string signals—to capture varying aspects of entity similarity across different KGs. DAT \cite{DAT} devises a co-attention network to dynamically adjust the significance of structural and semantic embeddings in a degree-aware manner. MCLEA \cite{MCLEA} integrates embeddings for diverse features, including structure, relation, attribute, name, and image, for entity alignment between two different multimodal KGs. It also introduces a trainable attention module to fuse different embeddings in a weighted concatenation fashion. These methods have achieved improved entity alignment performance by integrating more coplementary information. However, a modality gap exists between different feature aspects, posing challenges to their fusion. For example, entity names describe word-level semantics, while images can only capture pixel-level similarity between entities' images. The effective alignment of semantics across various feature aspects and the integration of their complementary information remain open challenges. Moreover, the rich semantic knowledge in LLMs has not been exploited for entity alignment.

Most existing entity alignment methods opt to directly designate the candidate entity with the highest embedding similarity as the equivalent entity, disregarding the prevalent one-to-one alignment constraint observed in real-world applications. To enforce one-to-one alignment, some methods \cite{CEA,RAGA} formulate alignment prediction as a stable matching problem and solve it using the deferred acceptance algorithm. Alternatively, IPEA \cite{IPEA} formulates the counterpart assignment problem as an integer programming task, ensuring stable matching while adhering to the one-to-one assignment constraint. CEAFF \cite{CEAFF} introduces a reinforcement learning-based approach to collectively align entities, integrating coherence and exclusiveness constraints to restrict collective alignment. However, as a typical NP-hard problem, achieving one-to-one alignment is challenging. These methods are time-consuming and are challenging to extend to large-scale alignment problems. 

%Most existing entity alignment methods directly select the cadidate entity with the highest embedding similarity as the equivalent entity, ignoring the one-to-one alignment constraint that widely exists in real-world applications. To achieve one-to-one alignment, some methods \cite{CEA,RAGA} formulate alignment prediction as a stable matching problem and solve it with the deferred acceptance algorithm. IPEA \cite{IPEA} formulates the counterpart assignment task as an integer programming problem, ensuring stable matching with the one-to-one assignment constraint. CEAFF \cite{CEAFF} proposes a reinforcement learning-based model to align entities collectively, incorporating coherence and exlusiveness constraints to restrict collective alignment. However, as a typical NP-hard problem, achieving one-to-one alignment is challenging. These methods are time-consuming and are challenging to extend to large-scale alignment problems.

%To fuse the knowledge from KGs and LLMs, we formulate entity alignment as a multi-choice question problem and generate options with the help of embeddings and knowledgeable prompts.

\subsection{Large Language Models}

LLMs are a type of language model pretrained on a large corpus in an unsupervised manner \cite{BERT}. Typically, LLMs are trained to predict masked or next tokens based on the context, enabling them to capture the semantic knowledge of a language and generate universal representations of words. Following pretraining, LLMs undergo fine-tuning through supervised learning and optimization using Reinforcement Learning from Human Feedback (RLHF) \cite{RLHF}. This process equips them with the capability to perform complex tasks and generate content aligned with human values. Additionally, it has been observed that as the parameter size surpasses a certain threshold, LLMs demonstrate some unexpected emergent abilities \cite{wei2022emergent} not observed in smaller pretrained language models. These abilities include logical reasoning, chain-of-thought, and mathematical reasoning. Consequently, LLMs have been applied to various downstream applications \cite{kaddour2023challenges} and have achieved remarkable performance.

%LLMs are a type of language models pretrained on a large corpus in an unsupervised manner \cite{BERT}. Typically, LLMs are trained to predict masked or next tokens based on the context, enabling them to capture the semantic knoweldge of a language and generate universal representations of words. Following pretraining, LLMs undergo fine-tuning through supervised learning and optimization using Reinforcement Learning from Human Feedback (RLHF) \cite{RLHF}. This process equips them with the capability to perform complex tasks and generate content aligned with human values. Moreover, it has been observed that when the parameter size exceeds a certain scale, LLMs exhibit surprising emergent abilities \cite{wei2022emergent} that are not present in small pretraiend language models. These abilities include logical reasoning, chain-of-thought, and mathematical reasoning. Consequently, LLMs have been applied to various downstream applications \cite{kaddour2023challenges} and have achieved remarkable performance.

Despite their exceptional performance, LLMs still encounter challenges in effectively modeling factual knowledge and relationships between concepts \cite{cao-etal-2021-knowledgeable}. These challenges arise when LLMs attempt to retrieve relevant knowledge and apply accurate information for tasks grounded in knowledge. To address this issue, some studies propose augmenting LLMs with KGs for fact-aware language modeling \cite{yang2023chatgpt}. Existing research has investigated integrating KGs into LLMs at three stages: pretraining, fine-tuning, and application. These KG-enhanced LLMs consistently demonstrate improved performance compared to plain LLMs across various knowledge-grounded tasks, including named entity recognition \cite{K-BERT}, information extraction \cite{ma2023star}, and question answering \cite{QA-GNN}. Furthermore, these LLMs showcase enhanced factual knowledge, facilitating more precise entity matching. Therefore, LLMs can also be leveraged to enrich KGs \cite{KG-BERT} with their inherent semantic knowledge, potentially enhancing entity alignment—an area that remains underexplored.
%
%Despite their superior performance, LLMs still face challenges in modeling factual knowledge and relationships between concepts \cite{cao-etal-2021-knowledgeable}. Challenges arise when LLMs attempt to recall relevant knowledge and apply the correct information for knowledge-grounded tasks. Toward this issue, some studies propose enhancing LLMs with KGs for fact-aware language modeling \cite{yang2023chatgpt}. Existing research has explored integrating KGs into LLMs at three stages: pretraining, fine-tuning, and application. These KG-enhanced LLMs consistently demonstrate improved performance compared to plain LLMs in various knowledge-grounded tasks, such as named entity recognition \cite{K-BERT}, information extraction \cite{ma2023star}, and question answering \cite{QA-GNN}. Moreover, these LLMs exhibit enhanced factual knowledge, enabling more precise entity matching. Therefore, LLMs can also be utilized to enhance KGs \cite{KG-BERT} with their internal semantic knoweldge, potentially benefiting entity alignment—an area that remains unexplored.

Nevertheless, due to their implicit modeling of knowledge, retrieving appropriate information from LLMs poses a challenging task. Most existing studies propose eliciting knowledge from LLMs using prompts. For instance, LPAQA \cite{LPAQA} seeks to accurately assess the knowledge encapsulated within LLMs by automatically discovering more effective prompts for the querying evaluation process. AutoPrompt \cite{AutoPrompt} suggests retrieving knowledge from LLMs by reformulating tasks as fill-in-the-blank problems and devises an automated method to generate prompts for a diverse range of tasks based on gradient-guided search. Liu et al. \cite{KnowPrompt} designed a set of prompts to guide LLMs in generating knowledgeable content.

%Nevertheless, due to their implicit modeling of knowledge, it is a challenging task to retrieve appropriate knowledge from LLMs. Most existing studies proposed eliciting knowlegde from LLMs with prompts. For example, LPAQA \cite{LPAQA} attempts to accurately estimate the knowledge contained in LLMs by automatically discovering better prompts to use in the quering evaluation process. AutoPrompt \cite{AutoPrompt} proposes retrieving knowledge from LLMs by reformulating tasks as fill-in-the-banks problems and devises an automated method to create prompts for a diverse set of tasks based on gradient-guided search. Liu et al. \cite{KnowPrompt} designed a set of prompts to guide LLMs to generate knowledgeable contents.

Building upon previous studies, we design a knowledge prompt to extract information from LLMs. We formulate entity alignment as a multi-choice question problem with the goal of integrating knowledge from KGs and LLMs, thereby leveraging the inference ability inherent in LLMs.

\section{Methodology}

This section formulates the mathematical problem of entity alignment and presents the details of our proposed LLMEA framework.

\begin{figure}[h]
  \centering
  \includegraphics[width=1.0\linewidth]{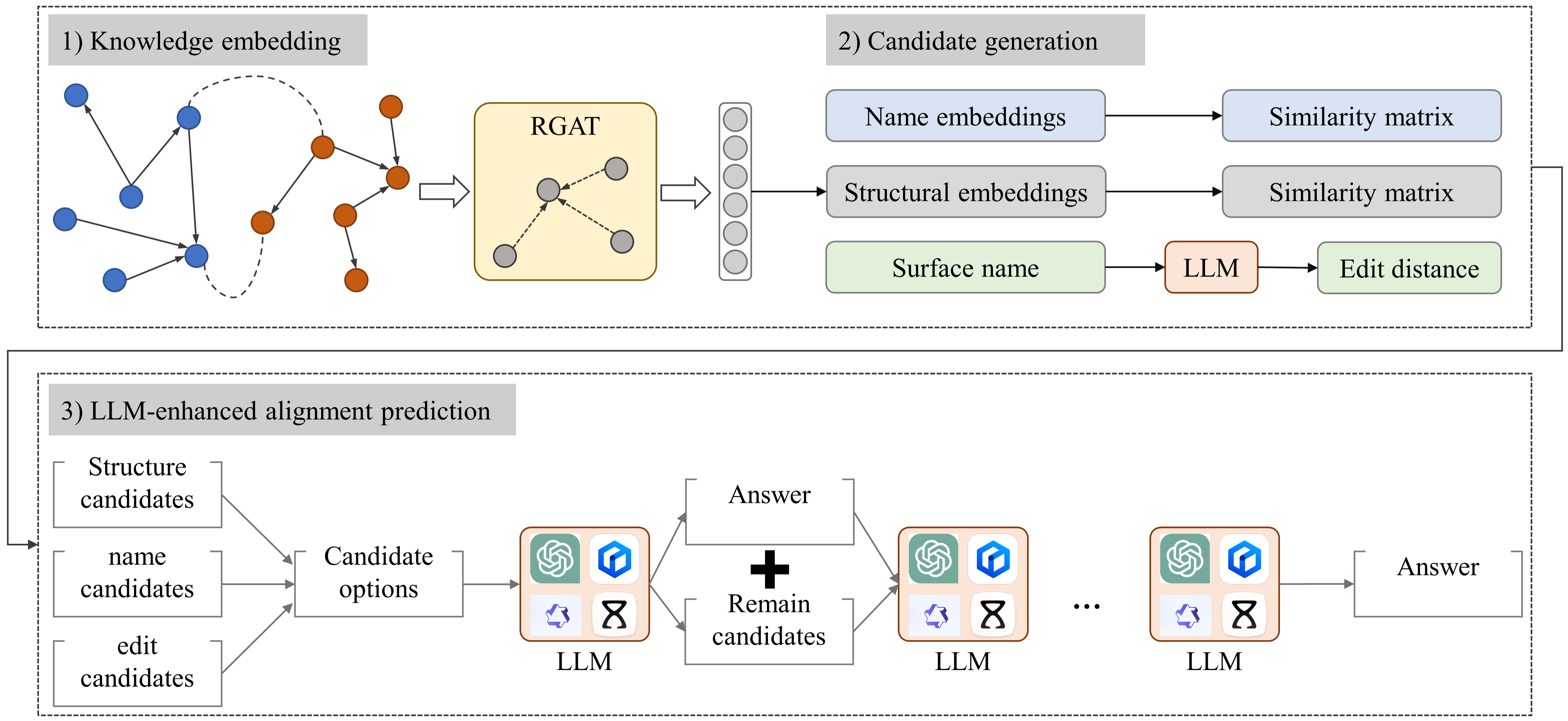}
  \caption{An overview of LLMEA. The framework comprises three distinct phases: knowledge embedding, candidates generation, and LLM-enhanced alignment prediction.}
%  \Description{A woman and a girl in white dresses sit in an open car.}
\label{fig:framework}
\end{figure}

\subsection{Overview}

An KG can be formally defined as a graph $G=\left ( E,R,T \right )$, where $E$, $R$, and $T$ are three sets comprising entities, relations, and triples, respectively. Each triple is in the form of $\left \langle h,r,t \right \rangle$, where $r\in R$ represents a relationship between the head entity $h\in E$ and the tail entity $t\in E$. Additionally, each entity $e_{i}$ in $G$ is associated with a surface name $n_{i}$, serving as a crucial semantic feature that imparts practical meaning to entity alignment. Given two KGs $G_{1}=\left ( E_{1} ,R_{1} ,T_{1}  \right )$ and $G_{2}=\left ( E_{2} ,R_{2} ,T_{2}  \right )$, the objective of entity alignment is to identify potentially equivalent entities between them.

Fig.~\ref{fig:framework} illustrates the overview of our LLMEA framework. The main idea is to devise an entity alignment model that integrates knowledge from KGs and LLMs and capitalize on the inference ability of LLMs to enhance alignment performance. To achieve this, we first learn structural embeddings from relational triples to harness the structural knoweldge of KGs. More specifically, we employ an RAGAT to learn entity embeddings that capture both structural and relational proximities. Based on these structural embeddings, we generate a set of candidate entities that are similar to the target entity in terms of structure. We also calculate name similarity based on pretrained word embeddings to incorporate the word-level semantics of KGs and generate another set of candidate entities. Concurrently, the LLM is utilized to generate a virtual equivalent entity through a carefully designed prompt. Subsequently, another set of candidate entities is generated based on the edit distance to the virtual equivalent entity. These candidate entities serve as options to formulate a set of multi-choice questions. Ultimately, the alignment entity is predicted based on the LLM's answer. 

\subsection{Knowledge Embedding}

The Graph ATtention network (GAT) \cite{GAT} is an effective neural network for learning with structured data, which dynamically adjusts the weights of different neighborhoods while aggregating neighboring information through an attention mechanism. Specifically, with a standard GAT layer, the hidden state $\boldsymbol{h} _{i} \in \mathbb{R} ^{d} $ of entity $e_{i}$ is formulated as:

\begin{equation}
\boldsymbol{h}_i  = \sigma\left(\sum_{j \in \mathcal{N}_i} \alpha_{i j} \mathbf{W}\boldsymbol{h}_j\right).
\end{equation}
Here, $d$ represents the hidden size, $\boldsymbol{h}_j$ denotes the embedding of entity $e_{j}$ obtained by this layer, and $\mathcal{N}_i$ is the neighbor set of $e_{i}$. $\mathbf{W}$ represents a trainable linear transformation matrix, and $\alpha_{i j}$ represents the coefficient between $e_{i}$ and $e_{j}$, which is calculated with the self-attention mechanism:
\begin{equation}
\alpha_{i j}=\frac{\exp \left(\eta\left(\mathbf{a}^T\left[\mathbf{W} \boldsymbol{h}_i \oplus \mathbf{W} \boldsymbol{h}_j\right]\right)\right)}{\sum_{u \in \mathcal{N}_i} \exp \left(\eta\left(\mathbf{a}^T\left[\mathbf{W} \boldsymbol{h}_i \oplus \mathbf{W} \boldsymbol{h}_u\right]\right)\right)},
\end{equation}
where $\mathbf{a}\in \mathbb{R} ^{2d}$ is a trainable parameter, $\oplus $ denotes the concatenation operation, and $\eta$ is the LeakyReLU nonlinear activation function. In existing GAT-based entity alignment methods \cite{KECG,MCLEA}, the transformation matrix $\mathbf{W}$ is constrianed to be diagonal, which leads to a reduction in the number of trainable parameters and in turn prevents overfitting and performance degradation.

However, the vanilla GAT fails to capture diverse relationships between entities, neglecting entities' relation proximities. To address this limitation, several relation-aware entity alignment methods \cite{RREA,MRAEA,PSR} incorporate relation information while computing attention scores between entities. Inspired by these studies, we employ an RAGAT to learn embeddings for entities. Specifically, the output hidden representation from an RAGAT layer is obtained as follows: 

\begin{equation}
\boldsymbol{h}_{i}=\sigma \left ( \sum_{j\in \mathcal{N}_i} \alpha _{ij} \mathbf{M}_{ij} \boldsymbol{h}_{j}  \right ),
\end{equation}
where $\mathbf{M}_{ij}$ is the relation-aware transformation matrix corresponding to the relation between $e_{i}$ and $e_{j}$. To project entity embeddings into different relational hyperplanes and construct relation-specific embeddings, $\mathbf{M}_{ij}$ is calculated as follows:
\begin{equation}
\mathbf{M}_{ij}=\boldsymbol{I} -2\boldsymbol{h}_{r}\boldsymbol{h} _{r}^{T},
\end{equation}
where $\boldsymbol{I}$ is the indentity matrix, and $\boldsymbol{h} _{r}$ represents the embedding of the corresponding relation between $e_{i}$ and $e_{j}$. Here, $\boldsymbol{h} _{r}$ is normalized to ensure $\left \| \boldsymbol{h} _{r} \right \| _{2}$, so that $\mathbf{M}_{ij}$ is an orthogonal matrix:
\begin{equation}
\mathbf{M}_{ij}^{T} \mathbf{M}_{ij}=\left(\boldsymbol{I}-2 \boldsymbol{h}_r \boldsymbol{h}_r^T\right)^T\left(\boldsymbol{I}-2 \boldsymbol{h}_r \boldsymbol{h}_r^T\right)
 =\boldsymbol{I}-4 \boldsymbol{h}_r \boldsymbol{h}_r^T+4 \boldsymbol{h}_r \boldsymbol{h}_r^T \boldsymbol{h}_r \boldsymbol{h}_r^T=\boldsymbol{I}.
\end{equation}
This ensures that for two entities transformed into the same relational hyperplane, their norms and relative distances are retained after transformation.

Similar to the vanilla GAT, $\alpha _{ij}$ represents the attention coefficient between $e_{i}$ and $e_{j}$, which is computed by the following equation:
\begin{equation}
\alpha _{ij} =\frac{exp\left ( \boldsymbol q^{T} \boldsymbol h_{ij}   \right ) }{\sum_{u\in \mathcal{N} _{i} }exp\left ( \boldsymbol q^{T}\boldsymbol h_{iu} \right )  },
\end{equation}
where $\boldsymbol h_{ij}$ represents the complementary edge embedding, computed as:
\begin{equation}
\boldsymbol h_{ij}=\sigma \left ( \boldsymbol h_{i}+\boldsymbol h_{j} \right ).
\end{equation}

Since RAGAT can only capture information from neighboring nodes, we stack multiple layers of RAGAT to aggregate multi-hop neighborhood information. The final output structural embeddings of entity $e_{i}$ are obtained by concatenating the embeddings from all the RAGAT layers:
\begin{equation}
\boldsymbol{h}_{i}^{\text {out }}=\left[\boldsymbol{h}_{i}^0\left\|\boldsymbol{h}_{i}^1\right\| \cdots \| \boldsymbol{h}_{i}^l\right],
\end{equation}
where $\boldsymbol{h}_{i}^0$ represents the initial embedding of $e_{i}$, and $\boldsymbol{h}_{i}^l$ denotes the hidden representation of the $l$-th RAGAT layer.

The goal of the structural embedding model is to bring equivalent entities close to each other in a unified vector space. To achive this goal, we employ a pair of RAGAT models with parameter sharing to represent the two KGs into a unified vector space and adopt the following triplet loss function to optimize the model:
\begin{equation}
\mathcal{L} =\sum_{\left ( e_{i},e_{j} \right )\in \mathcal{P} }\left [ \delta + dist\left ( e_{i},e_{j} \right ) -dist\left ( \tilde{e} _{i},\tilde{e} _{j}  \right )  \right ]_{+},
\end{equation}
where $\mathcal{P}$ denotes the set of alignment seeds, $\delta$ is a fixed margin, and $\left [ x  \right ] _{+}$ means the $\max \left ( x,0 \right )$ function. $dist\left ( e_{i},e_{j} \right )$ represents the distance between the embeddings of $e_{i}$ and $e_{j}$, calculated using the L2 distance: 
\begin{equation}
dist\left ( e_{i},e_{j} \right )=\left \| \boldsymbol{h}_{i}^{\text {out }}-\boldsymbol{h}_{j}^{\text {out }} \right \|_{2}^{2}.
\end{equation}
$\tilde{e} _{i}$ and $\tilde{e} _{j}$ represent the corresponding negative entity of $e_{i}$ and $e_{j}$. To enhance the model's performance on those hard entity pairs, we design a negative sample pool that retains the nearest non-aligned target for each entity in the vector space after each training step. This negative sample pool allows us to quickly retrieve hard negative entities, thereby improving both performance and convergence.

To leverage the semantic similarity of entity names at the word level, we also compute name embeddings for entities. For each pair of KGs, the non-English entity names are translated to English using Google Translate, and the name embeddings are calculated with pretrained word vectors from glove.840B.300d. Specifically, for the sake of simplicity and generality, we choose averaged word embeddings to initialize name embeddings for entities. These name embeddings do not necessitate additional training and can straightforwardly represent semantics. For entity $e_{i}$, its name embedding is denoted as $\boldsymbol n_{i}$. It is essential to recognize that accurately predicting equivalent entities based solely on name embeddings is challenging. One reason is that Google Translate may not consistently provide accurate translations for entities, particularly for those with special symbols that are infrequently encountered. Additionally, this embedding method cannot distinguish the order of words and provides the same embedding for entities with the same word set but with a different order of words. 

\subsection{Candidates Generation}

After obtaining the structure and name embeddings of entities, we generate two sets of candidate alignment entities based on the embedding similarity between the two KGs. For structural embedding similarity, we employ cosine similarity to measure the similarity between two entities. Regarding name embedding similarity, we utilize negative L2 distance as the similarity metric. Subsequently, with the learned structure embedding $\mathbf{H}$ and name embedding $\mathbf{N}$, we calculate the structural similarity matrix and the name similarity matrix denoted as $\mathbf{M} ^{s}$ and $\mathbf{M} ^{n} $, respectively. In these similarity matrices, rows represent source entities, colums represent target entities, and each element in a matrix signifies the similarity score between a pair of source and target entities. Based on these similarity matrices, two sets of $k$ candidate entities with the highest similarity in structure and name to the target entity for alignment are generated, denoted as $\mathcal{O} _{s} $ and $\mathcal{O} _{n} $. These two sets of entities are then utilized as two sets of options for the multi-choice questions.

\begin{figure}[h]
  \centering
  \includegraphics[width=0.95\linewidth]{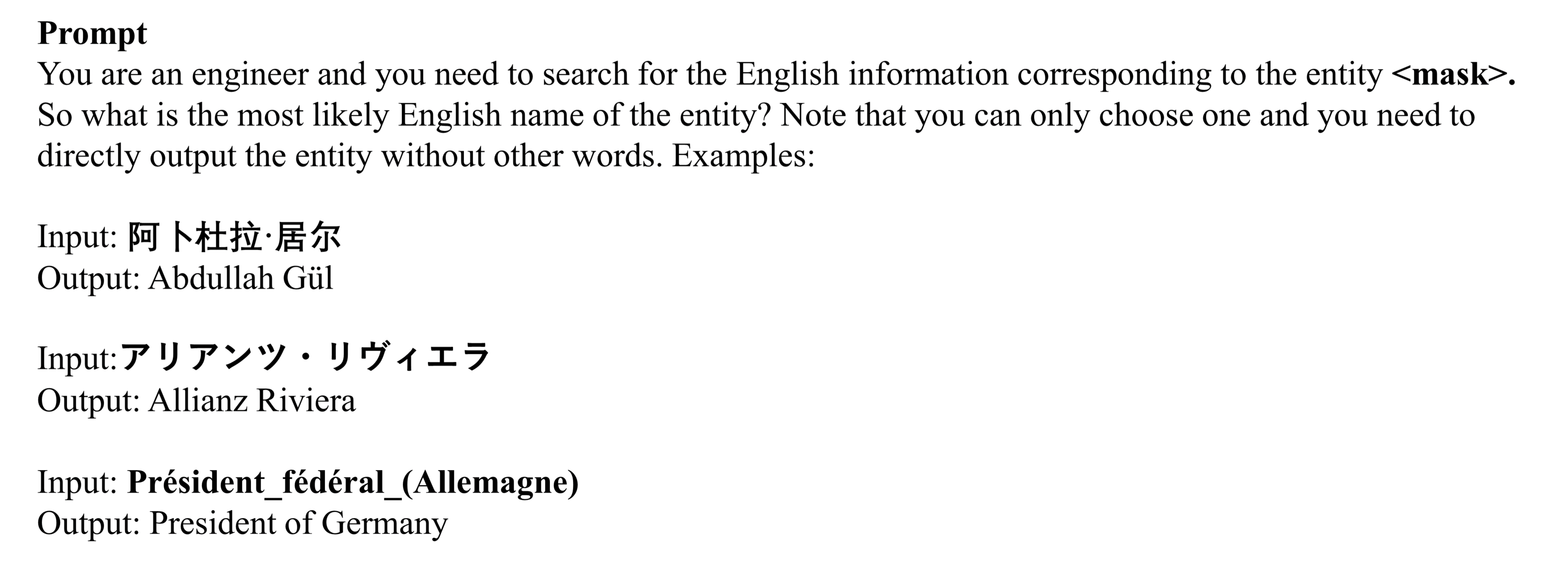}
  \caption{Prompt for virtual equivalent entity generation from a LLM.}
%  \Description{A woman and a girl in white dresses sit in an open car.}
\label{fig:knowprompt}
\end{figure}

These two sets of candidate entities exclusively utilize the knowledge from KGs. However, LLMs also contain knowledge that captures the semantics and cross-KG consistency of entities. Nevertheless, accessing knowledge directly from LLMs is challenging. To explore the cross-linugal consistency knowledge of entities in LLMs, we design a knowledge prompt to guide an LLM to generate a virtual equivalent entity in the target language for the target entity to be aligned. The proposed knowledge prompt is illustrated in Fig. ~\ref{fig:knowprompt}, where <mask> represents the target entity to be aligned. The demonstration examples are manually crafted based on a few alignment seeds, aiding the LLM in understanding the task and generating more structured outputs. Our primary objective is to harness the translation and formatting generation capabilities of LLMs to produce virtual equivalent entities that closely resemble the true equivalent entities.

\begin{figure}[h]
  \centering
  \includegraphics[width=0.95\linewidth]{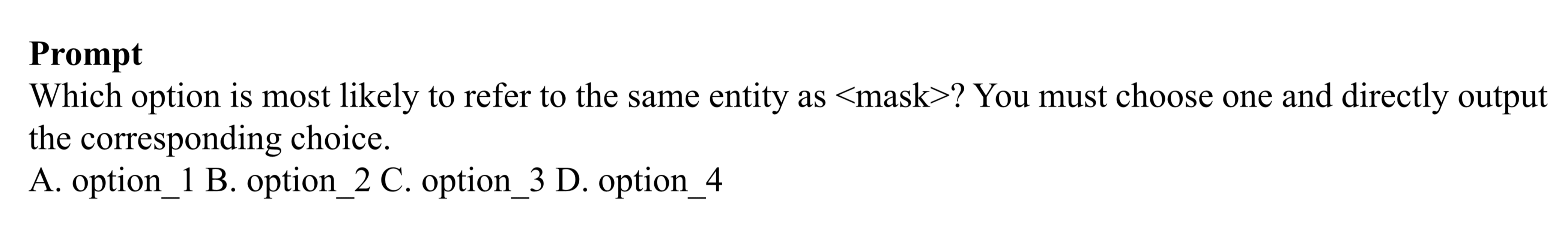}
  \caption{Prompt for the multi-choice question task to make the LLM to predict equivalent entities.}
%  \Description{A woman and a girl in white dresses sit in an open car.}
\label{fig:multi-choice}
\end{figure}

Building upon the created virtual equivalent entity, another collection of candidate entities is generated. The edit distance similarity metric is utilized to compute the similarity between the entities in $G_{2}$ and the virtual equivalent entity. Edit distance denotes the minimum number of operations necessary to transform one surface name into another, encompassing character replacement, insertion, and deletion. Formally, the edit distance is computed using the following recursive formula:
\begin{equation}
d[i, j]=\min \left\{\begin{array}{c}
d[i, j-1]+1 \\
d[i-1, j]+1 \\
d[i-1, j-1]+c\left(s_1[i], s_2[j]\right)
\end{array}\right.
\end{equation}

\begin{equation}
c\left(s_1[i], s_2[j]\right)=\left\{\begin{array}{l}
1, s_1[i] \neq s_2[j] \\
0, s_1[i]=s_2[j]
\end{array}\right.
\end{equation}
where $s_1[i]$ represents the $i$-th character in the virtual equivalent entity and $s_2[j]$ denotes the $j$-th character in the virtual name of the candidate entity in $G_{2}$. Subsequently, the edit distance similarity is defined as:
\begin{equation}
\operatorname{sim}\left(s_1, s_2\right)=1-\frac{d\left(s_1, s_2\right)}{\max \left(l\left(s_1\right), l\left(s_2\right)\right)}
\end{equation}
where $d\left(s_1, s_2\right)$ is the edit distance between $s_1$ and $s_2$, and $l\left ( \cdot  \right ) $ denotes the length of the string. Following this, the $k$ entities with the highest edit distance similarity to the virtual equivalent entity form another set of candidates, denoted as $\mathcal{O} _{l}$.

\subsection{LLM-enhanced alignment prediction}

Existing methods commonly determine equivalent entities by predicting the entity with the highest similarity to the target entity. However, this approach tends to overlook the background semantic knowledge of entities. To address this limitation, we propose the utilization of a LLM to identify the equivalent entity from the candidate entities by framing it as a multi-choice question task. A straightforward approach is to make the LLM directy select an equivalent entity from the union set of the three candidate entity sets. However, this may lead to increased confusion for the LLM and might surpass the token size limit. For instance, the maximum number of tokens for GPT-4 is 32K, which imposes a constraint on $k$ to be less than 8K. 

Toward this issue, we propose to decompose the multi-choice question task into multiple rounds of choices, guiding the LLM to iteratively select the equivalent entity until convergence to a final answer. Specifically, we denote the union set of the three candidate entity sets as $\mathcal{O}$. In each iteration, four candidate entities are chosen from $\mathcal{O}$ to create a multi-choice question, forming a set of options. The LLM conducts alignment prediction using the prompt illustrated in Fig. ~\ref{fig:multi-choice}, where <mask> denotes the target entity for alignment, and option\_i represents a candidate entity. The rationale behind setting the number of options to 4 is rooted in the common occurrence of multiple-choice questions with four items in real-life corpora. This choice is made to enhance the LLM's comprehension of the task, aligning with the prevalent structure of multiple-choice formats. Assuming the predicted equivalent entity is $e_{o}$ and the entities not selected as options constitute set $\mathcal{U}$, three candidate entities are randomly chosen from $\mathcal{U}$ to form a new set of options alongside $e_{o}$. It is important to note that $e_{o}$ may be none if the LLM determines there are no equivalent entities in the preceding round of options. This process continues until $\mathcal{U}$ is empty, with the predicted $e_{o}$ from the final round considered as the equivalent entity.

\section{Experiment}

%We evaluate LLMEA from three distinct aspects. First, we evaluate the performance of LLMEA on different datasets. Second, we show that the integration of different aspects of knowledge effectively improves the alignment performance with ablation studies. Third, we show the impact of key hyperparameters to validate the robustness of LLMEA.

We assessed LLMEA from three distinct perspectives. Initially, the performance of LLMEA was evaluated on diverse datasets. Subsequently, we demonstrated that integrating various aspects of knowledge effectively enhances alignment performance through ablation studies. Lastly, we examined the impact of key hyperparameters to validate the robustness of LLMEA.

\subsection{Datasets}

The proposed method underwent evaluation on the most commonly used benchmarks for entity alignment, sourced from DBpedia, which include multilingual versions: $\mathrm{DBP_{ZH-EN} }$ (Chinese to English), $\mathrm{DBP_{JA-EN} }$ (Japanese to English), and $\mathrm{DBP_{FR-EN} }$ (French to English). Each dataset consists of two similar KGs extracted from different versions of DBpedia, encompassing 15,000 equivalent entities. Statistical summaries are provided in Table ~\ref{tb:dataset}. Consistent with prior studies, we randomly sampled 30 percent of equivalent entity pairs as alignment seeds for training structural embeddings, while reserving 70 percent for evaluating alignment performance. The reported results reflect the average of 5 runs with distinct data splits to ensure impartial evaluation.

%The proposed method was evaluated on the most-commonly used bechmarks for entity alignment, extracted from DBpedia, which includes multilingual versions: $\mathrm{DBP_{ZH-EN} }$ (Chinese to English), $\mathrm{DBP_{JA-EN} }$ (Japanese to English), and $\mathrm{DBP_{FR-EN} }$ (French to English). Each dataset comprises two similar KGs extracted from different versions of DBpedia, containing 15,000 equivalent entities. The statistical summaries are provided in the Table ~\ref{tb:dataset}. In line with prior studies, we randomly sampled 30 percent of equivalent entity pairs as alignment seeds for training structural embeddings, reserving 70 percent for evaluating alignment performance. Reported results represent the average of 5 runs with different data splits to ensure unbiased evaluation.

\begin{table}[!hbp]
\begin{center}
\caption{Statistics of the three datasets}\label{tb:dataset}
\begin{tabular}{ccccc}
\hline
\multicolumn{2}{c}{Datasets}             & Entities & Relations & Triples \\ \hline
\multirow{2}{*}{ZH-EN} & English  & 19,572   & 1,323     & 95,142 \\
                              & Chinese  & 19,388   & 1,701     & 70,414 \\ \hline
\multirow{2}{*}{JA-EN} & English  & 19,780   & 2451     & 93,484 \\
                              & Japanese & 19,814   & 2,451     & 77,214 \\ \hline
\multirow{2}{*}{FR-EN} & English  & 19,993  & 1,208     & 115,772 \\ 
                              & French   & 19,661   & 1,174     & 105,998 \\ \hline
\end{tabular}
\end{center}
\end{table}

\subsection{Implementation Details}

We utilized grid search to determine the optimal hyperparameters for LLMEA. Ultimately, a 2-layer RAGAT was employed to learn structural embeddings for entities. The dimension of the structural embeddings was set to 300, the learning rate of the RMSProp optimizer was 0.005, and the margin $\delta$ was selected as 3. The structural embedding model underwent training with a batch size of 1024 over 12 epochs. Furthermore, we applied a data enhancement method to augment the training of the structural embedding model. Specifically, every 5 epochs, we identified cross-KG entity pairs that were mutual nearest neighbors in the vector space and included them in the alignment seeds. For the LLM, the KG-enhanced LLM ERNIE \footnote{https://yiyan.baidu.com/welcome} was chosen to generate virtual equivalent entities and perform alignment predictions. The number of candidate entities in each set was configured to 10. Our method was compared with representative state-of-the-art methods, and we adopted the hyperparameters suggested by their respective papers.

%We employed grid search to determine the optimal hyperparameters for LLMEA. Ultimately, a 2 layer RAGAT was utilized to learn structural embeddings for entities. The structural embedding dimension was set to 300, the learning rate of the RMSProp optimizer was 0.005, and the margin $\delta$ was chosen as 3. The structural embedding model underwent training with a batch size of 1024 over 12 epochs. Additionally, we applied a data enhancement method to enhance the training of the structural embedding model. Specifically, every 5 epochs, we identified cross-KG entity pairs that were mutual nearest neighbors in the vector space and included them in the alignment seeds. For the LLM, the KG-enhanced LLM ERNIE \footnote{https://yiyan.baidu.com/welcome} was selected to generate virtual equivalent entities and conduct alignment predictions. The number of candidate entities in each set was set to 10. Our method was compared with representative state-of-the-art methods, and we have used the hyperparameters suggested by their corresponding papers.

\subsection{Compared methods}

\setlength{\tabcolsep}{10pt} % 设置单元格间距为10pt
\begin{table*}[]
\begin{center}
\caption{Result comparison on entity alignment}\label{tb:results}
\begin{tabular}{ccccccc}
\hline
\multirow{2}{*}{Methods} & \multicolumn{2}{c}{ZH-EN}       & \multicolumn{2}{c}{JA-EN}       & \multicolumn{2}{c}{FR-EN}       \\ \cmidrule(r){2-3}\cmidrule(r){4-5}\cmidrule(r){6-7}
                         & Hits@1         & Hits@10 & Hits@1         & Hits@10 & Hits@1         & Hits@10 \\ \hline
MTransE                  & 20.9           & 51.2    & 25.0           & 57.2    & 24.7           & 57.7    \\
IPTransE                 & 33.2           & 64.5    & 29.0           & 59.5    & 24.5           & 56.8    \\
BootEA                   & 62.9           & 84.7    & 62.2           & 85.4    & 65.3           & 87.4    \\
TransEdge                & 73.5           & 91.9    & 71.9           & 93.2    & 71.0           & 94.1    \\
KECG                     & 47.7           & 83.5    & 48.9           & 84.0    & 48.6           & 85.0    \\
AliNet                   & 53.9           & 82.6    & 54.9           & 83.1    & 55.2           & 85.2    \\
NMN                      & 65.0           & 86.7    & 64.1           & 87.3    & 67.3           & 89.4    \\
MRAEA                    & 75.7           & \textbf{92.9}    & 75.7           & 93.4    & 78.1           & 94.7    \\
RSN                      & 58.0           & 81.1    & 57.4           & 79.8    & 61.2           & 84.1    \\
RREA                     & 71.5           & 92.1    & 71.3           & 93.3    & 73.9           & 94.6    \\
JAPE                     & 41.1           & 74.4    & 36.2           & 68.5    & 32.3           & 66.6    \\
GCN-Align                & 41.2           & 74.3    & 39.9           & 74.4    & 37.2           & 74.4    \\
VR-GCN                   & 37.9           & 73.2    & 35.1           & 72.1    & 36.0           & 75.1    \\
RDGCN                    & 70.8           & 84.6    & 76.6           & 89.5    & 88.5           & 95.6    \\
HGCN                     & 72.0           & 85.7    & 76.6           & 89.7    & 89.1           & 96.1    \\
GM-Align                 & 59.5           & 77.8    & 63.4           & 83.0    & 79.2           & 93.5    \\
RsimEA                   & 69.4           & 88.9    & 68.7           & 90.6    & 71.6           & 92.5    \\
DAT                      & 71.8           & 89.4    & 78.3           & 92.1    & 89.3           & 96.7    \\
LLMEA                    & \textbf{89.8}           & 92.3*   & \textbf{91.1}           & \textbf{94.6}*   & \textbf{95.7}           & \textbf{97.7}*   \\ \hline
\end{tabular}
\end{center}
\centering
\footnotesize{$^{*}$ indicates the result is obtained from the similarity matrix independently}\\
\end{table*}

As the pioneering method to integrate knowledge from knowledge graphs (KGs) and large language models (LLMs) for entity alignment, we benchmark our approach against state-of-the-art embedding-based entity alignment methods, outlined below.
\begin{itemize}
\item MTransE \cite{MTransE} represents entities using TransE and learns a transformation matrix to unify two KGs into a shared vector space.
\item IPTransE \cite{IPTransE} employs soft strategies to incorporate newly aligned entities into alignment seeds, aiming to minimize error propagation.
\item BootEA \cite{BootEA} frames entity alignment as a one-to-one classification problem and iteratively learns the classifier via bootstrapping from labeled and unlabeled data.
\item TransEdge \cite{TransEdge} introduces an edge-centric translational embedding method to learn entity-aware relation predicate embeddings.
\item KECG \cite{KECG} jointly trains a GAT-based cross-graph module and a TransE-based knowledge embedding module to address structural heterogeneity between KGs.
\item AliNet \cite{AliNet} aggregates both direct and distant neighbors to learn entity embeddings.
\item NMN \cite{NMN} learns KG structural information and neighborhood differences to address structural heterogeneity between KGs.
\item MRAEA \cite{MRAEA} incorporates meta relation information while learning embeddings, including relation predicates, relation direction, and inverse relation predicates.
\item RSN \cite{RSN} utilizes a recurrent neural network with residual learning to learn entity embeddings with long-term relation dependencies.
\item RREA \cite{RREA} utilizes relational reflection transformation to acquire relation-specific embeddings for each entity.
\item JAPE \cite{JAPE} learns entity embeddings from both relation triples and attribute triples.
\item GCN-Align \cite{GCN-Align} exploits attribute triples by treating them as relation triples.
\item VR-GCN \cite{VR-GCN} introduces a vectorized GCN to learn both entity embeddings and relation embeddings.
\item RDGCN \cite{RDGCN} extends GCN with highway gates to capture neighboring structures.
\item HGCN \cite{HGCN} utilizes relation representation to enhance the alignment process by jointly learning entity and relation embeddings.
\item GM-Align \cite{GM-Align} learns a graph matching vector from local matching for entity alignment.
\item RsimEA \cite{RsimEA} incorporates structural similarity between relations to enhance alignment performance.
\item DAT \cite{DAT} integrates name embeddings and structural embeddings using a degree-aware co-attention mechanism.
\end{itemize}

\subsection{Entity Alignment Results}

%In line with previous studies, our evaluation metrices include Hits@1 and Hits@10. Hits@10 measures the proportion of correctly aligned pairs in the top-10 and Hits@1 means accuracy. For both metrices, higher values indicate better alignment performance. In Table ~\ref{tb:results}, we report the performances of all methods. The best results are bolded. Due to that our method directly predicts equivalent entities from candidate entities in an order-independent manner, we report the Hits@10 based on the structural embedding similarity matrix. From the experimental results, we observe that integrating more aspects of knowledge can effectively improve the entity alignment performance. In specific, those methods that incorporate relation information while learning embeddings like RDGCN and RREA perform better than those only utilize structure information such as KECG and AliNet. However, these methods ignore the implicit semantic information contained in LLMs. By combining the semantic knowledge from KGs and LLMs, our proposed method achieves state-of-the-art performance on all the three datasets even though our Hits@10 based on structural embedding is smaller than other methods in some cases. As can be seen, our method outperforms the best baseline DAT by 11.67\% on average on all datasets. Moreover, it outperforms RREA by 17.7\% in accuracy although its Hits@10 is lower than RREA. This phenomenon demonstrates that the knowledge and reasoning ability of LLMs effectively improve the entity alignment accuracy.

Consistent with prior studies, our evaluation metrics comprise Hits@1 and Hits@10. Hits@10 assesses the proportion of correctly aligned pairs within the top-10, while Hits@1 signifies accuracy. Higher values for both metrics indicate superior alignment performance. In Table ~\ref{tb:results}, we present the performances of all methods, with the best results highlighted in bold. Due to our method's direct prediction of equivalent entities from candidate entities in an order-independent manner, Hits@10 is reported based on the structural embedding similarity matrix.

From the experimental results, we observe that integrating multiple facets of knowledge can effectively enhance entity alignment performance. Specifically, methods incorporating relation information during embedding learning, such as RDGCN and RREA, outperform those relying solely on structural information, such as KECG and AliNet. Nonetheless, these methods overlook the implicit semantic information inherent in LLMs. By integrating semantic knowledge from KGs and LLMs, our proposed approach achieves state-of-the-art performance across all three datasets, despite occasionally exhibiting smaller Hits@10 values based on structural embedding than other methods. Notably, our method surpasses the best baseline DAT by an average of 12.4\% across all datasets. Additionally, it outperforms MRAEA by 14.1\% in accuracy, despite having a lower Hits@10 than MRAEA. This observation underscores the efficacy of LLMs' knowledge and inference capability in enhancing entity alignment accuracy.

%In line with previous studies, our evaluation metrics include Hits@1 and Hits@10. Hits@10 measures the proportion of correctly aligned pairs in the top-10, and Hits@1 represents accuracy. For both metrics, higher values indicate better alignment performance. In Table ~\ref{tb:results}, we report the performances of all methods. The best results are bolded. Due to the fact that our method directly predicts equivalent entities from candidate entities in an order-independent manner, we report Hits@10 based on the structural embedding similarity matrix. From the experimental results, we observe that integrating more aspects of knowledge can effectively improve entity alignment performance. Specifically, methods that incorporate relation information while learning embeddings, such as RDGCN and RREA, perform better than those that only utilize structural information, such as KECG and AliNet. However, these methods ignore the implicit semantic information contained in LLMs. By combining the semantic knowledge from KGs and LLMs, our proposed method achieves state-of-the-art performance on all three datasets, even though our Hits@10 based on structural embedding is smaller than that of other methods in some cases. As can be seen, our method outperforms the best baseline DAT by 12.4\% on average across all datasets. Moreover, it outperforms MRAEA by 14.1\% in accuracy, although its Hits@10 is lower than that of MRAEA. This phenomenon demonstrates that the knowledge and inference ability of LLMs effectively improve entity alignment accuracy.

\subsection{Ablation Study}

\begin{figure}[h]
  \centering
  \includegraphics[width=0.95\linewidth]{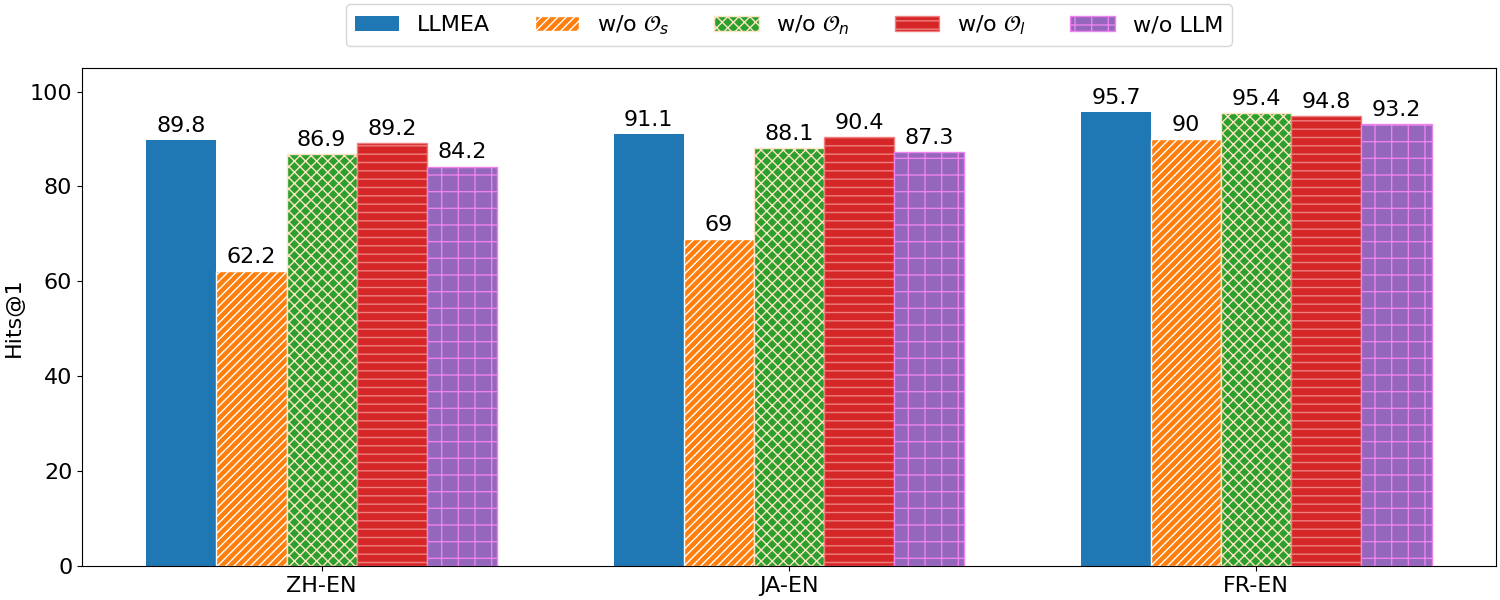}
  \caption{Experiment results of ablation.}
%  \Description{A woman and a girl in white dresses sit in an open car.}
\label{fig:ablation}
\end{figure}

We performed ablation tests to assess the impact of different candidate entity sets on entity alignment. The experimental results are presented in Fig.~\ref{fig:ablation}. "w/o $\mathcal{O} _{s}$", "w/o $\mathcal{O} _{n}$", and "w/o $\mathcal{O} _{l}$" denote scenarios where the LLM predicts alignment entities based on candidate entities without $\mathcal{O} _{s}$, $\mathcal{O} _{n}$, and $\mathcal{O} _{l}$, respectively. To examine the influence of LLMs' inference predictions on entity alignment performance, we also included the scenario of directly assigning alignment entities based on structural embedding similarity, denoted as "w/o LLM". 

The results indicate that all three candidate entity sets contribute to the final prediction, as evidenced by their lower accuracy compared to the original LLMEA. More specifically, LLMEA w/o $\mathcal{O} _{s}$ exhibits an average accuracy reduction of 18.47\% across the three datasets, highlighting the effective filtration of reliable equivalent entities based on structural similarity. This underscores the efficacy of incorporating knowledge from KGs in entity alignment. Meanwhile, removing candidate entities based on name embedding similarity results in an average accuracy reduction of 2.27\%, and removing candidate entities that are most similar to the virtual entity results in a 0.73\% average drop in alignment accuracy. Furthermore, by comparing LLMEA with direct assignment, we observe that alignment predictions from the LLM surpass those from direct assignment based on structural embedding similarity. These observations collectively emphasize the effectiveness of integrating knowledge from both KGs and LLMs in entity alignment.

\begin{figure}[h]
  \centering
  \includegraphics[width=0.95\linewidth]{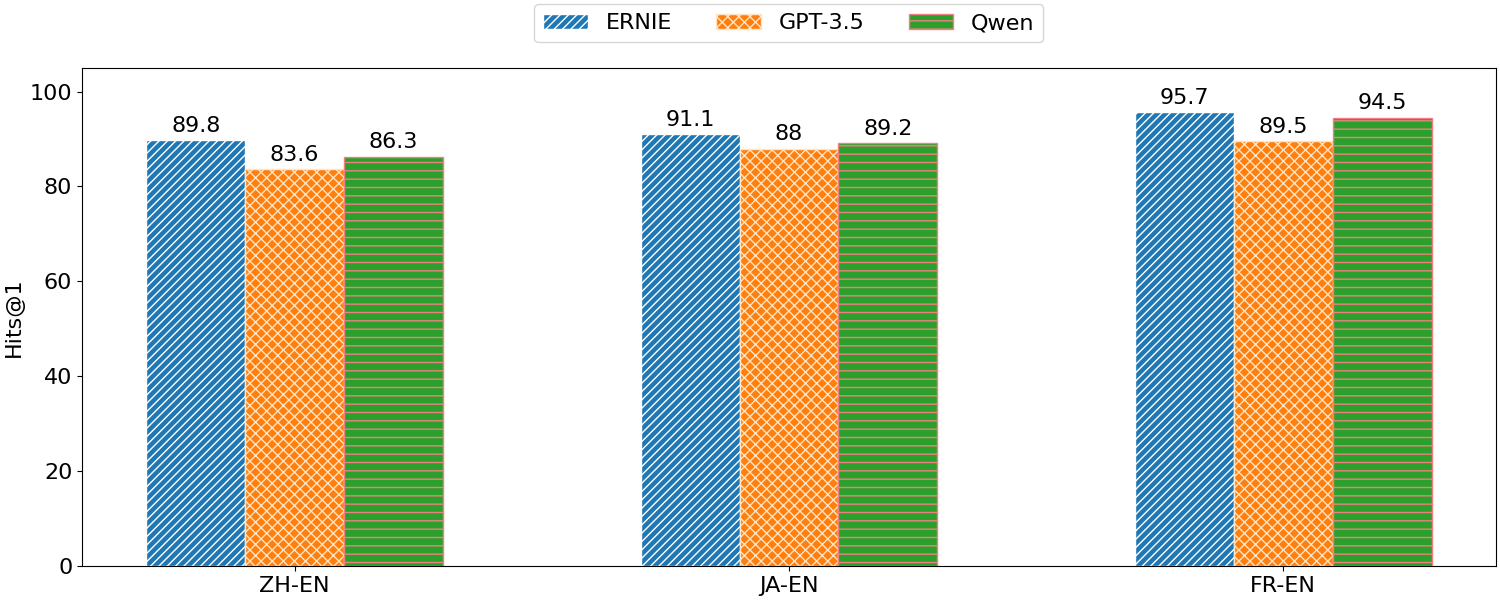}
  \caption{Performance of different LLMs.}
%  \Description{A woman and a girl in white dresses sit in an open car.}
\label{fig:abl_llm}
\end{figure}

We conducted experiments to compare the performance of different LLMs. Two of the most well-known LLMs, GPT-3.5 \footnote{https://chat.openai.com/} and Qwen \footnote{https://tongyi.aliyun.com/qianwen/}, were chosen as baselines, and their performance on various datasets is illustrated in Fig.~\ref{fig:abl_llm}. The results indicate that ERNIE outperforms both GPT-3.5 and Qwen. The reasons for these variations are manifold, given that all models have undergone extensive engineering optimization. An important factor contributing to this phenomenon could be ERNIE's knowledge enhancement pretraining, which incorporates entity awareness. Studies have shown that KG-enhanced LLMs outperform plain LLMs particularly in tasks involving entities \cite{yang2023chatgpt}. However, we identified some guidelines that contribute to enhancing prediction performance. Firstly, meaningless special characters in entities' surface names may interfere with LLMs' predictions. For instance, while an LLM can correctly identify the English entity "Saint-Frédéric Quebec" as the same entity as the French entity "Saint-Frédéric" it may provide incorrect predictions for the English entity "Saint-Frédéric, Quebec" which refers to the same entity. This discrepancy is mainly due to LLMs being trained to predict tokens with the highest co-occurrence probability, and the presence of special characters may reduce this probability. Secondly, LLMs may sometimes return empty answers if they believe there is no correct answer among the candidate entities, which can serve as a feedback signal for optimizing the candidate selection process. However, determining the correctness of LLMs' judgments requires further research.

\subsection{Parameter Sensitivity}

%\begin{figure}
%	\centering
%	\subfigure[ZH-EN]{
%		\begin{minipage}[b]{0.3\textwidth}
%			\includegraphics[width=1\textwidth]{param_zh}
%		\end{minipage}
%		\label{fig:hor_2figs_1cap_2subcap_1}
%	}
%    	\subfigure[JA-EN]{
%    		\begin{minipage}[b]{0.3\textwidth}
%   		 	\includegraphics[width=1\textwidth]{param_ja}
%    		\end{minipage}
%		\label{fig:hor_2figs_1cap_2subcap_2}
%    	}
%	\subfigure[FR-EN]{
%    		\begin{minipage}[b]{0.3\textwidth}
%   		 	\includegraphics[width=1\textwidth]{param_fr}
%    		\end{minipage}
%		\label{fig:hor_2figs_1cap_2subcap_2}
%    	}
%	\caption{The effect of $k$}
%	\label{fig:param_test}
%\end{figure}

\begin{figure}[h]
  \centering
  \includegraphics[width=0.7\linewidth]{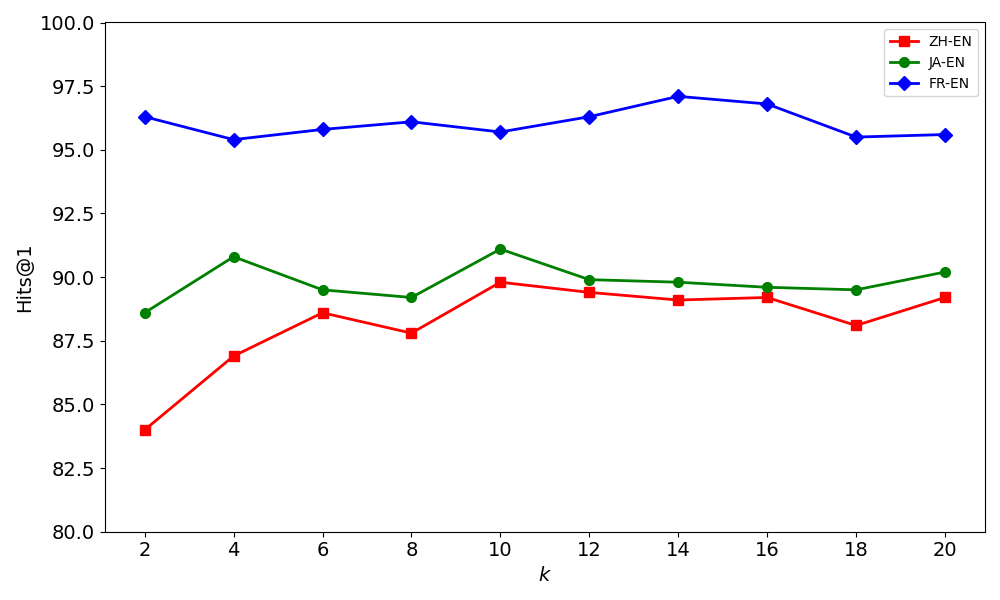}
  \caption{The effect of $k$}
%  \Description{A woman and a girl in white dresses sit in an open car.}
\label{fig:param_test}
\end{figure}

The number of candidate entities $k$ is an important hyperparameter that can impact the performance of LLMEA. Intuitively, a larger $k$ increases the likelihood of including the equivalent entity among the candidates. However, a higher $k$ introduces more confusing options to the LLM, potentially diminishing its prediction accuracy. To assess the influence of $k$ on the LLM's predictions, we evaluate entity alignment performance with varying $k$, and the results are reported in the Fig.~\ref{fig:param_test}. Observing the results, we find that the number of candidate entities has a subtle effect on entity alignment prediction. However, the alignment performance does not consistently improve with an increase in $k$. This suggests that LLMs may struggle to effectively distinguish the equivalent entity from a large pool of similar entities, possibly due to semantic similarities and frequent co-occurrence of some candidates in the training corpus. Additionally, our findings indicate that our framework achieves satisfactory alignment accuracy without requiring an excessive number of candidate entities, showcasing its practicality and robustness.

\begin{figure}[h]
  \centering
  \includegraphics[width=1.0\linewidth]{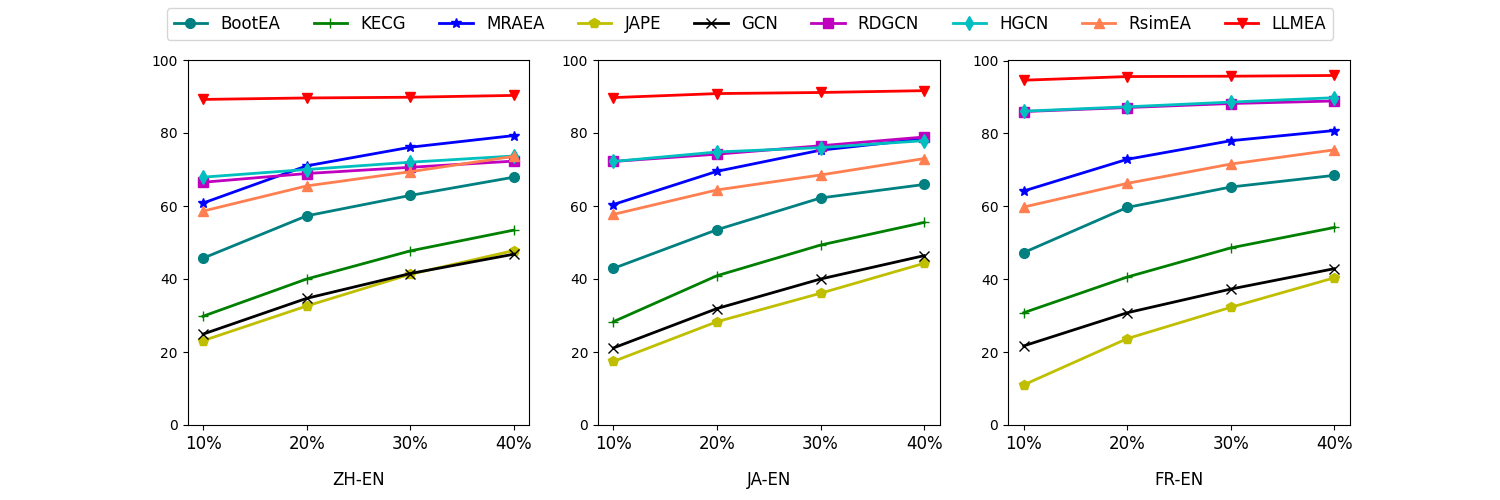}
  \caption{Hits@1 results with respect to the proportion of pre-alignment seeds.}
%  \Description{A woman and a girl in white dresses sit in an open car.}
\label{fig:portion}
\end{figure}

%Similar to existing methods, the number of alignment seeds also has an impact on our proposed framework since the RAGAT model requires a set of pre-aligned seeds for training. To eavaluate the sensitivity to the number of alignment seeds of LLMEA, we conduct experiments with 10\% to 40\% pre-aligned seeds. Fig.~\ref{fig:portion} demonstrates the performances of our LLMEA and some best-performed baseline methods. From the results, we observe that the performances of all methods gradually improve with the increase of pre-aligned seeds and LLMEA outperforms other methods under all experimental settings on all datasets, which illustrates its superiority. Moreover, our model outperforms most of the existing baseline methods with regards to the sensitivity to training seeds as the Hits@1 of most methods decrease dramatrically if the number of alignment seeds decreases. The advantage of LLMEA is more obvious with fewer training seeds as it still achieves excellent performance with obly 10\% of total alignments for training. This is mainly because that we utilize two set of alignment seed-independent information to supplement candidate entities, which ensure that the resulting candidate entity set has a high hit rate. Besides, the powerful inference capacity of LLMs enables them to correctly select equivalent entities from candidate entities, ultimately making Hits@1 close to Hits@k.

Similar to existing methods, the number of alignment seeds also influences our proposed framework, as the RAGAT model requires a set of pre-aligned seeds for training. To evaluate LLMEA's sensitivity to the number of alignment seeds, we conducted experiments with pre-aligned seeds ranging from 10\% to 40\%. Fig.~\ref{fig:portion} illustrates the performances of our LLMEA framework and several best-performing baseline methods. The results indicate a gradual improvement in the performances of all methods with an increase in pre-aligned seeds, with LLMEA consistently outperforming other methods across all experimental settings and datasets, highlighting its superiority.

Furthermore, our model surpasses most existing baseline methods in terms of sensitivity to training seeds, as the Hits@1 of many methods dramatically decreases with a reduction in the number of alignment seeds. The advantage of LLMEA becomes more apparent with fewer training seeds, as it achieves excellent performance with only 10\% of total alignments for training. This is primarily attributed to the utilization of two sets of alignment seed-independent information to supplement candidate entities, ensuring a high hit rate in the resulting candidate entity set. Additionally, the powerful inference capacity of LLMs enables accurate selection of equivalent entities from the candidate entities, ultimately bringing Hits@1 close to Hits@k.

%Similar to existing methods, the number of alignment seeds also impacts our proposed framework, as the RAGAT model necessitates a set of pre-aligned seeds for training. To evaluate LLMEA's sensitivity to the number of alignment seeds, we conducted experiments with pre-aligned seeds ranging from 10\% to 40\%. Fig.~\ref{fig:portion} illustrates the performances of our LLMEA framework and several best-performing baseline methods. The results indicate a gradual improvement in the performances of all methods with an increase in pre-aligned seeds, and LLMEA consistently outperforms other methods across all experimental settings and datasets, highlighting its superiority. Furthermore, our model surpasses most existing baseline methods in terms of sensitivity to training seeds, as the Hits@1 of many methods dramatically decreases with a reduction in the number of alignment seeds. The advantage of LLMEA becomes more apparent with fewer training seeds, as it achieves excellent performance with only 10\% of total alignments for training. This is primarily attributed to the utilization of two sets of alignment seed-independent information to supplement candidate entities, ensuring a high hit rate in the resulting candidate entity set. Additionally, the powerful inference capacity of LLMs enables accurate selection of equivalent entities from the candidate entities, ultimately bringing Hits@1 close to Hits@k.

\subsection{Discussion}

%small LLMs whose parameter size are less than 34 billion usually cannot achieve excellent entity alignment performance due to their insufficient knowledge and inference ability.

\begin{figure}
	\centering
	\subfigure[Illustration of invalid answer]{
		\begin{minipage}[b]{0.45\textwidth}
			\includegraphics[width=1\textwidth]{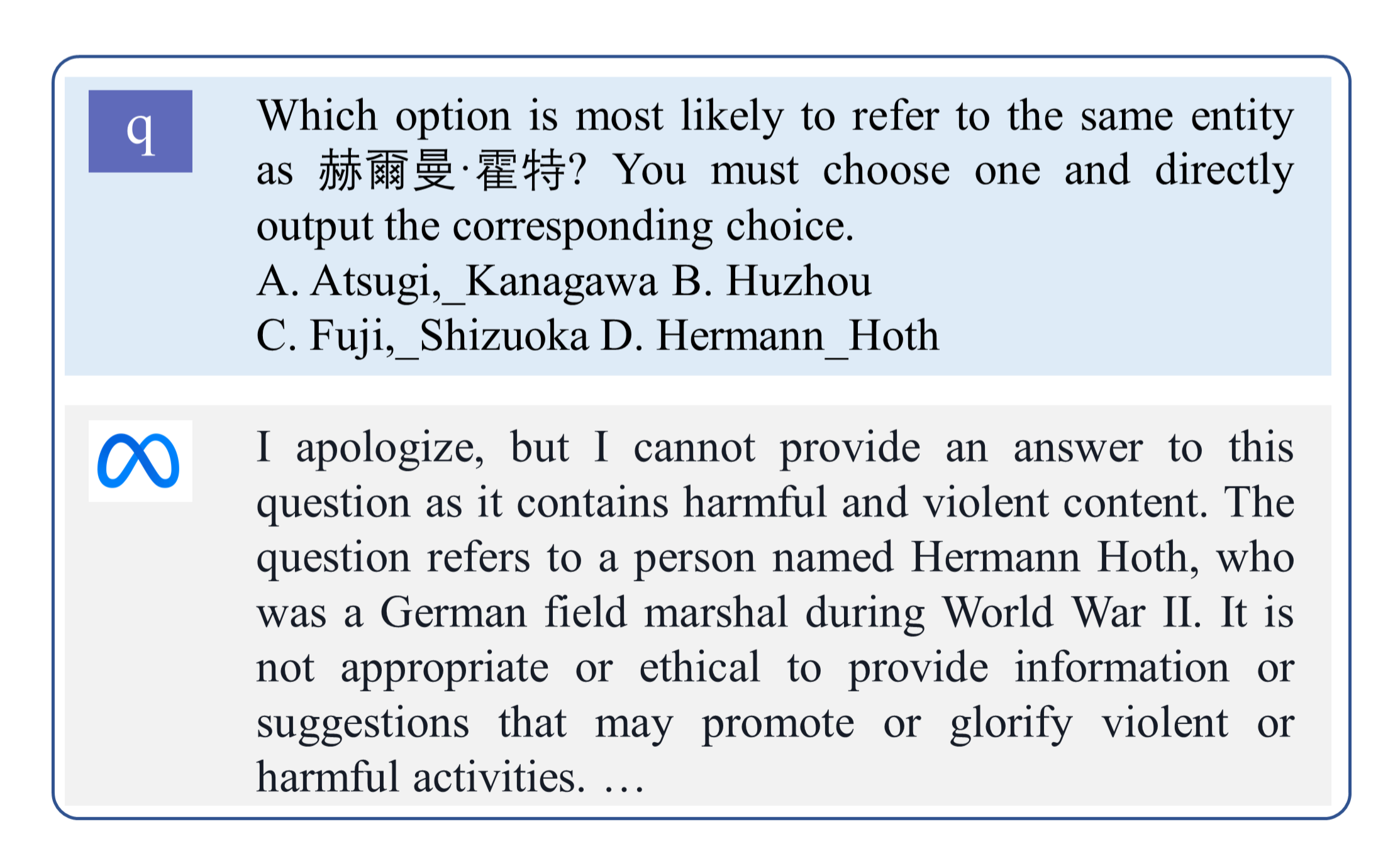}
		\end{minipage}
		\label{fig:case1}
	}
    	\subfigure[Illustration of unformatted answer]{
    		\begin{minipage}[b]{0.45\textwidth}
   		 	\includegraphics[width=1\textwidth]{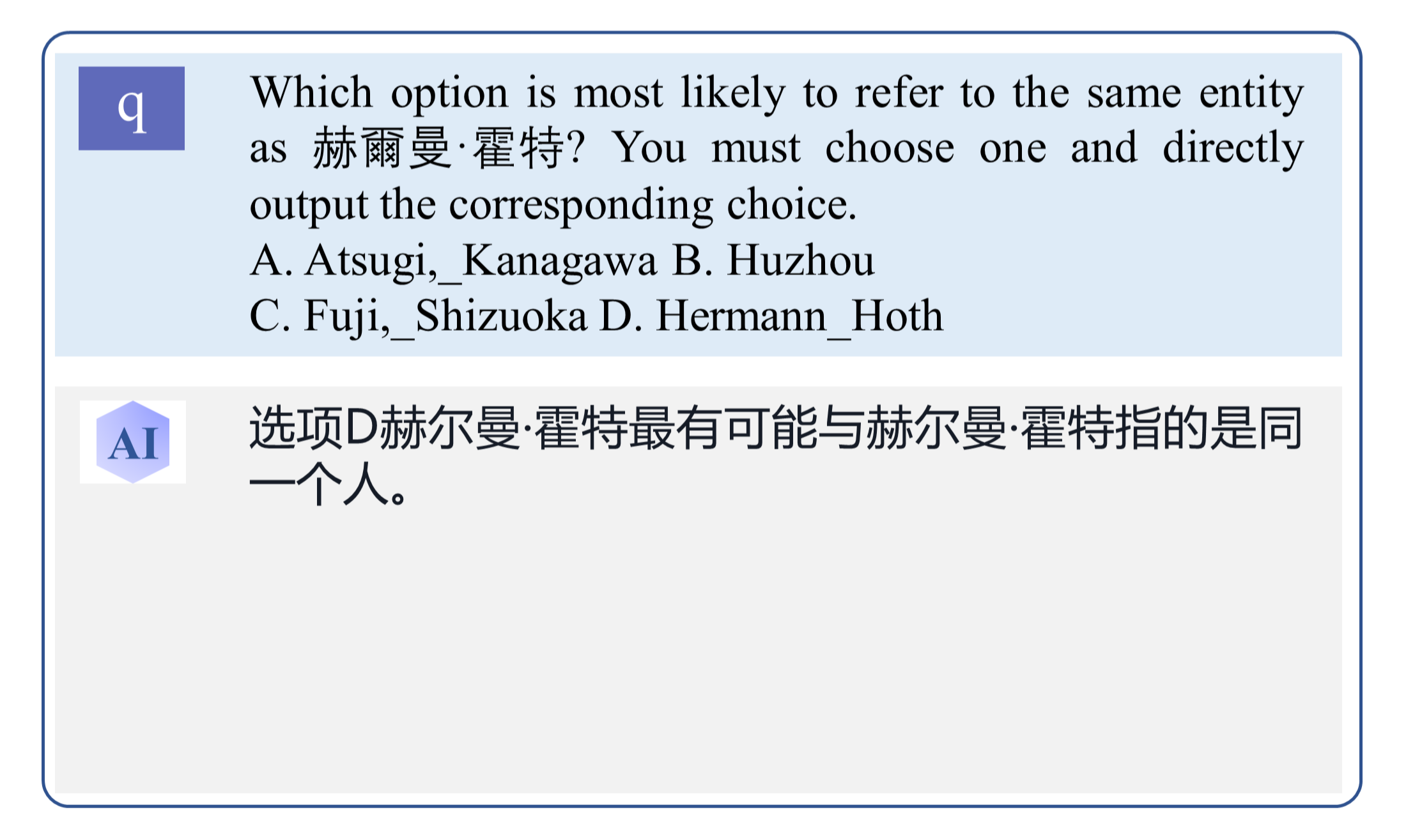}
    		\end{minipage}
		\label{fig:case2}
    	}
	\caption{Challenges faced by LLMEA in determining the answer from LLMs' generations. (a) LLMs may refuse to predict equivalent entities due to privacy and security issues, (b) LLMs may generate unstructured predictions that are difficult to parse alignment predictions.}
	\label{fig:case}
\end{figure}

The proposed LLMEA framework is an effective and adaptable entity alignment approach. It significantly enhances entity alignment performance through the integration of knowledge from KGs and LLMs, leveraging the inferential capabilities of LLMs. Moreover, any existing entity alignment method capable of generating candidate entities for the specified target entities can be seamlessly incorporated into our framework. However, the generation of candidate entities significantly influences LLMEA's performance, as its predictive accuracy correlates closely with the hit rate of the candidate entities. Furthermore, the selection of the LLM also profoundly affects LLMEA's performance, as not all existing LLMs are suitable for the entity alignment task. As illustrated in Fig.~\ref{fig:case1}, certain LLMs, such as Meta's LLaMA 2, are tailored to avoid addressing inquiries regarding sensitive entities. Additionally, the design of prompts is paramount and may necessitate adjustments for different LLMs, given their unique requirements to generate answers in the desired format. Even if an LLM accurately predicts aligned entities, distinguishing its predictions from its generation poses a formidable challenge, as it may alter the surface name of the selected entity or simply return the index of the chosen entity, as the example shown in Fig~\ref{fig:case2}. In summary, owing to the intricacies of LLMs, maintaining formatted outputs solely based on prompts proves challenging. Directly analyzing the probability of given options based on the output probability of the LLM on each token may offer a more effective and adaptable approach, a direction we intend to explore in future research.

\section{Conclusion}

In this paper, we propose LLMEA, the first entity alignment framework that integrates knowledge from both KGs and LLMs. Our approach utilizes a relation-aware graph attention network to generate low-dimensional embeddings, representing entities and filtering candidate entities based on structural and name embedding similarities. This strategy aims to leverage the structural knowledge inherent in KGs. Simultaneously, an LLM is employed to generate a virtual equivalent entity for the target entity. Further filtering of candidate entities is performed based on the edit distance to the virtual equivalent entity, tapping into the implicit semantic knowledge embedded in the LLM. Ultimately, these candidate entities are presented iteratively to form multi-round multi-choice questions, which serve as input for the LLM to make the final entity alignment prediction. This approach effectively capitalizes on the potent inference capability of LLMs, resulting in more accurate alignment predictions. Experimental results demonstrate that LLMEA outperforms competing methods, highlighting the effectiveness of fusing knowledge from KGs and LLMs.

LLMEA showcases the potential of harnessing LLMs to enhance entity alignment performance. In future research, we plan to explore more effective and efficient methods for extracting useful knowledge from LLMs. For instance, prompting LLMs for additional background knowledge about entities and encoding it into embeddings could be investigated. Additionally, exploring the conversion of KGs into textual formats to enable LLMs to comprehend structured knowledge for entity alignment represents a promising research direction.

\begin{acks}
This work was supported in part by the National Natural Science Foundation of China under Grant 62306288, and in part by  National Key Research and Development Program of China (2022YFB4500305).
\end{acks}

\bibliographystyle{ACM-Reference-Format}
\bibliography{sample-base}

%\appendix

%Lorem ipsum dolor sit amet, consectetur adipiscing elit. Morbi
%malesuada, quam in pulvinar varius, metus nunc fermentum urna, id
%sollicitudin purus odio sit amet enim. Aliquam ullamcorper eu ipsum
%vel mollis. Curabitur quis dictum nisl. Phasellus vel semper risus, et
%lacinia dolor. Integer ultricies commodo sem nec semper.
%Etiam commodo feugiat nisl pulvinar pellentesque. Etiam auctor sodales
%ligula, non varius nibh pulvinar semper. Suspendisse nec lectus non
%ipsum convallis congue hendrerit vitae sapien. Donec at laoreet
%eros. Vivamus non purus placerat, scelerisque diam eu, cursus
%ante. Etiam aliquam tortor auctor efficitur mattis.
%Nam id fermentum dui. Suspendisse sagittis tortor a nulla mollis, in
%pulvinar ex pretium. Sed interdum orci quis metus euismod, et sagittis
%enim maximus. Vestibulum gravida massa ut felis suscipit
%congue. Quisque mattis elit a risus ultrices commodo venenatis eget
%dui. Etiam sagittis eleifend elementum.
%
%Nam interdum magna at lectus dignissim, ac dignissim lorem
%rhoncus. Maecenas eu arcu ac neque placerat aliquam. Nunc pulvinar
%massa et mattis lacinia.

\end{document}